# Teaching a Language Model to Distinguish Between Similar Details using a Small Adversarial Training Set

Chris Achard

*The University of Texas at Austin*

## Abstract

Language models can achieve high accuracy on natural language tasks such as NLI, but performance suffers on manually created adversarial examples. We investigate the performance of a language model trained on the Stanford Natural Language Inference (SNLI) corpus on a manually created adversarial test set. We then improve the model's performance by fine tuning the model on a small, manually created adversarial *training* set, designed to help the language model to learn to differentiate between similar words and phrases in the data. We show an increase in accuracy on the adversarial test set **(+ 13%)** while still maintaining good performance on the original NLI task. We also show an **increase in accuracy from 91.2% to 92.9%** on the most similar contradictions in the SNLI test set (as judged by cosine similarity).

## 1. Introduction

As language models become more and more common in our daily lives, it becomes more and more important to understand the limits of those models, and how simple changes to test and training sets can significantly change the accuracy of such models. One way to understand the limitations of language models is to evaluate them on a common benchmark task such as NLI.

Natural Language Inference (NLI), is the task of predicting whether two sentences (a "premise" and a "hypothesis") have a relationship that is either *entailed, neutral,* or a *contradiction* (MacCartney and Manning 2008). The Stanford Natural Language Inference (SNLI) corpus is a set of 570k such sentence pairs, broken into training, test, and validation sets (Bowman et al. 2015). The SNLI corpus can be used to train a wide variety of language models for the NLI task.

One way to reduce the accuracy of a language model is to test it on a very specific, human created adversarial test set. An adversarial set is specifically created to take advantage of one or more weaknesses detected in the model (Robin Jia and Percy Liang. 2017).

For this study, we train a language model and match the state of the art performance for the NLI task as of 2017 (89.2%). Then we demonstrate how a simply constructed adversarial test set can easily reduce that accuracy, which may indicate that the language model isn't learning a very robust representation of the data. After investigating exactly how that adversarial test set breaks the model, we attempt to fix this model deficiency by fine-tuning on a manually constructed adversarial *training* set, which is designed to help increase the model accuracy on the adversarial test set, while not degrading its performance on the original NLI task.

## 2. Approach

### 2.1 Baseline Model

The language model we use throughout this paper is the ELECTRA Small model (Clark et al., 2020). We used the starter code provided for the UT NLP 2023 class (https://github.com/gregdurrett/fp-dataset-artifacts) and trained it on the SNLI training set for 3 epochs. This achieved a baseline SNLI test set accuracy of **89.2%**, which matches state-of-the-art results as of 2017, (although the current state of the art is 93.1%) (Bowman et al., 2015).

We will use this baseline model and accuracy result throughout this analysis to help determine the performance against various test sets and the efficacy of our proposed fixes.

## 2.2 Adversarial Test (Challenge) Set

Previous works have created adversarial challenge sets either by paraphrasing parts of the input (Madnani and Dorr, 2010) or by concatenating inputs in a way that leaves the answer unchanged (Jia and Liang, 2017). These types of simple automatic adversarial sample generation techniques are difficult for the NLI task however, as there is no general way to assure the new samples do not change the label.

Instead, we first focus on manual adversarial challenge set creation. This is similar to work which creates manual "contrast sets" (Gardner et al., 2020), but instead of creating samples which change the label, we focus on modifying existing samples in a way which keeps the label constant, but which the human believes will be more difficult for the model to interpret.

To create the challenge set, the human first gains experience by trying different modifications on various sentences, and tries to "trick" the model (by changing the model prediction without changing the true gold label between the premise and hypothesis). After categorizing certain types of errors that the model makes, the human then creates a new adversarial test set (challenge set) without model feedback, and that is what we used for our analysis. The various types of errors the human tried to take advantage of are detailed in the next main section. This is similar to the work of Bartolo et al., because there is a model in the loop to start, but differs because our final adversarial test set is constructed with no model in the loop (Bartolo et al., 2020).

## 2.3 Approach for Fixing Model Shortcomings

After creating the adversarial test set, the same human then creates another set of adversarial examples to be used to finetune the model. The intended idea is to "inoculate" the model for the adversarial *test* set by using a similar, small adversarial *training* set (Liu et al., 2019).

### 3. Identifying Model Errors

After experimenting with the model to determine how to change either the premise or hypothesis sentence, we've identified **one primary** type of error with **three subtypes** that we can use to get the model to give the wrong label without changing the underlying meaning or relationship. To further focus our efforts and get more consistent results, we restricted these changes to sentences with the *contradiction* label. The two other labels (*entailment* and *neutral*) may have similar but distinct types of errors.

## 3.1 General Error Category: Shared Words and Ideas

For sentence pairs within the *contradiction* label, we identified that **shared words and ideas** between the premise and hypothesis could lead the model to incorrectly predict *entailment* or *neutral*.

To better quantify and visualize this across the entire test set, we calculated GloVe embeddings for every word in each sentence (excluding stop words), and averaged those word embeddings together, effectively creating a bag of words (BOW) embedding for each sentence individually (Pennington et al., 2014).

We then took the cosine similarity score between each premise and hypothesis, then graphed the percent of correct and incorrect answers over the cosine similarity scores from 0.5 to 1.0:

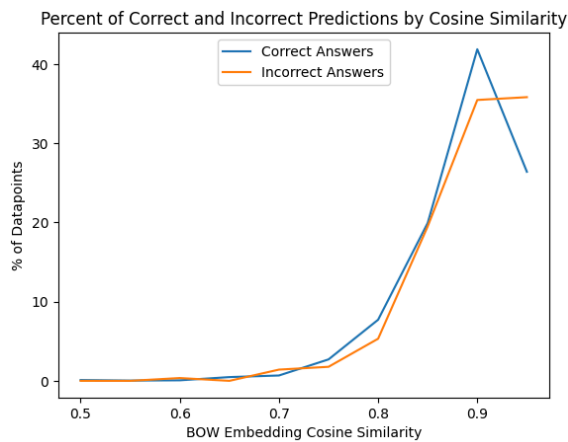

The plots for correct and incorrect answers differ slightly, but it is especially apparent at the high end of cosine similarity (where the BOW embeddings are very similar), as there is a higher percent of *correct* answers with *lower* cosine similarity, and a higher percent of *incorrect* errors at a *higher* cosine similarity. Specifically, this means that sentences in the test set that are **more similar** (have more words or synonyms in common), are **more difficult** for the model to predict correctly (increasing the error rate).

To see this more clearly, we can focus on just cosine similarity > 0.8, and look at the **cumulative %** of the correct and incorrect answers at that similarity **or higher:**

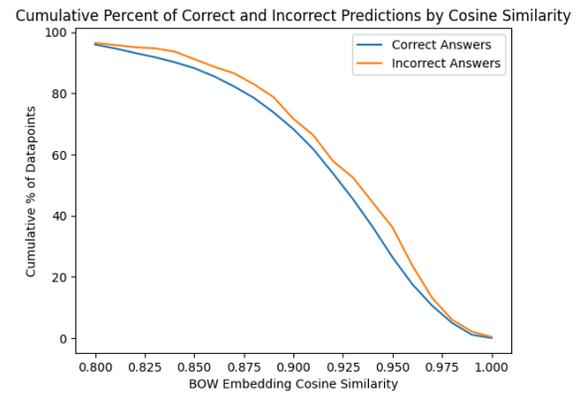

That graph now shows us that for every cosine similarity value above 0.8, there are a higher cumulative percent of incorrect answers than correct answers. That means that **incorrect answers have a higher chance of being similar than correct ones.** (Note that this does not apply to total counts, simply because there are many more correct predictions than incorrect ones. This effect is only shown when looking at the *percent* of correct and incorrect answers).

In our results we will show that with our model fix we are able to increase the accuracy of this set (contradictions with a cosine similarity of 0.8 or greater) from 91.2% to 92.9%, which further verifies that this is a real and fixable problem with the original model.

We can see this more clearly by looking at three different subtypes of this error category, with specific examples for each subcategory.

**3.2 Error Subcategory #1: Negation**

One way to get the model to predict *entailment* instead of *contradiction* is to take words or a phrase from one sentence, and express the opposite (negative) of that phrase in the other. For example, we took a sentence pair from the test dataset that the model got correct:

*Original Premise:* `"A skateboarder skates in the pool"`

***Original Hypothesis***: "A skate swims in the pool."
***Prediction:*** Contradiction **(correct)**

And added a negative statement to the premise which mirrored the verb of the hypothesis:

***New Premise:*** "A skateboarder skates in the pool **and doesn't swim**."
***New Hypothesis***: "A skate swims in the pool."
***Prediction:*** Entailment **(incorrect)**

Although the label should still be "*contradiction*", there are now too many similar words and phrases in both sentences for the model to properly distinguish the correct label.

**3.3 Error Subcategory #2: Abstract Detail**

Another way to add more similar words or phrases between the premise and hypothesis is to add them in an abstract, or more conceptual way. Because the SNLI dataset was created from image captions, it only handles *physical, real world observations* very well, and doesn't know much about abstract concepts of thinking, time frames like before and after, dreaming, wishing, etc. We can therefore trick the model by increasing the similarity of the premise and hypothesis by adding these abstract phrases:

***Original Premise:*** "Two men workers doing their job with their supervisor."
***Original Hypothesis***: "Two men are taking their lunch break."
***Prediction:*** Contradiction **(correct)**

but we can flip the prediction by modifying it to:

***New Premise:*** "Two men workers doing their job with their supervisor **but want to eat lunch**."
***New Hypothesis***: "Two men are taking their lunch break **with their supervisor**."
***Prediction:*** Entailment **(incorrect)**

In this example we modify both the premise and hypothesis to increase the word similarity, and include the idea of what the men in the image *want* to be doing. A human knows that *wanting* to do something is not the same as actually doing it (especially when it explicitly states they are not doing it earlier in the sentence), but the model is tricked because this abstract idea is more difficult based on the training dataset.

**3.4 Error Subcategory #2: Preposition Confusion**

The last common subcategory we identified is that the model gets confused if we switch the prepositions in a sentence so that the sentence means something different. For example:

***Original Premise: "**A worker is working outside of a restaurant.**"***
***Original Hypothesis***: "A worker is working in a restaurant."
***Prediction:*** Contradiction **(correct)**

Here the model correctly predicts the difference between "in" and "outside", but if we swap in a new preposition:

***New Premise:*** "A worker is working **on top of** a restaurant."
***New Hypothesis***: "A worker is working in a restaurant."
***Prediction:*** Entailment **(incorrect)**

The model can't distinguish between being *in* vs *on top of* the restaurant, but that would be clear to a human. (This commonly works for a wide variety of prepositions like *near, above* and *below*).

### 3.5 Creating the Adversarial Test and Training Sets

After identifying those common errors, we then manually created an adversarial test set of **100 test contradictions** by taking random contradiction samples from the SNLI test set, and modifying them in similar ways as described above.

We also created **100 new training samples** in the same way (except by sampling from the SNLI *training* set).

### 3.6 Initial Performance on the Adversarial Test Set

We then used the model to predict labels for our new adversarial test set. If the test set was perfectly similar to the SNLI test set, then we would expect a similar performance (89.2%), but since we have designed the test set adversarially (to specifically take advantage of the model's weaknesses), the result is much lower: **62%.**

It's also interesting to look at the distribution of the predicted labels on the new adversarial test set, which should show all contradictions, but shows that many contradictions are predicted to be entailments or neutral:

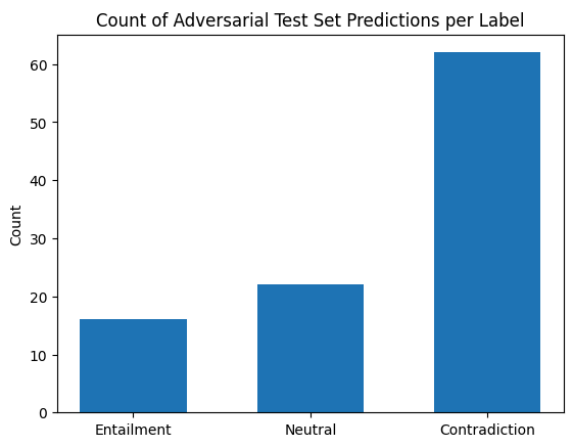

Our primary goal for the rest of the paper is to fix the incorrect predictions in the above graph (entailment and neutral predictions), which should all be labeled as contradictions.

### 4. Fixing the Model

To fix these errors in the baseline model, we used the 100 new training samples we created and fine-tuned the previously trained model. Because these training samples were only *contradiction* samples (since that is what we are investigating), we also added 100 samples for each other label: *entailment*, and *neutral* from the original SNLI training set to balance out the fine-tuning and not only bias the model towards contradiction.

We trained with those 300 samples for 3 additional epochs. We lowered the learning rate to 1e-5 because we found that higher learning rates started to degrade the model performance on the original SNLI test set. A further study could look more deeply at the training and finetuning methods, to see if there is a way to further fine-tune with the new training set without any degradation to the original test set performance.

### 5. Results

When trained for 3 epochs on the SNLI training set, our baseline ELECTRA-small model achieved an accuracy of **89.2%** across the SNLI test set. After finetuning on the new adversarial training set, the total accuracy dropped just a bit to **88.9%.** A drop in total accuracy is a potential weakness of this approach, but still a very good score for the SNLI dataset and task.

The real test is whether or not the new training set improved the scores of both the adversarial test set and of the most similar contradictions of the SNLI test set (BOW cosine similarity > 0.8). With the baseline that value was **62%** and after the finetuning fix that jumped to **75%**! That shows that the adversarial training set actually worked to improve the deficiencies we found and modeled in the adversarial test set.

We can also plot that on the bar graph showing the total count of adversarial test set predictions, to get a better feel for how the changes affect each label's prediction:

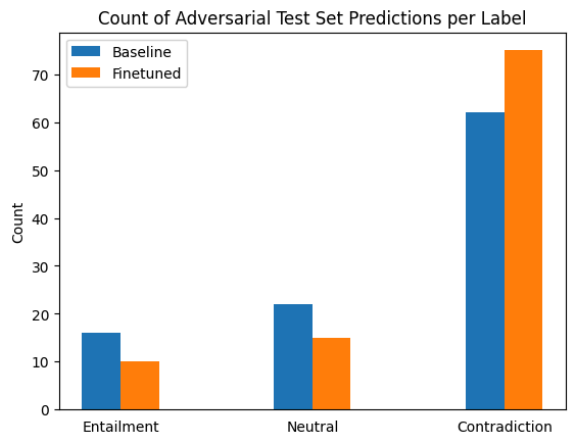

With that graph we can clearly see the increase in correct labels (*contradiction*), and a decrease in both of the incorrect labels (*entailment* and *neutral*).

The final overall statistic to look at is the accuracy of the most similar contradiction labels in the SNLI test set. With the baseline model, we were able to achieve **91.2%** accuracy, while the finetuned model achieved an accuracy of **92.9%** on the same set: an increase of **+ 1.7%.** This shows that the fix that worked for the adversarial test set also addressed the model deficiencies we found in the original SNLI test set.

### 5.1 Ablations

To get a better feel for how exactly the adversarial training set affects the accuracy of the adversarial test set, the most similar contradictions in the SNLI test set, as well as the entire SNLI test set, we can run a few ablation studies to look at the impact of the new training data used. This will also give us a much better feel for the types of tradeoffs we can make with this type of fix, and could point us towards future research directions.

First, we analyze the decision to finetune with 100 new contradictions (the adversarial training set), combined with 100 original SNLI training samples from each of the other labels. We do this by comparing the baseline model, a model finetuned with 300 original SNLI training samples., a model finetuned with *only* the 100 adversarial contradiction training samples, and finally a model finetuned with the 100 adversarial contradictions with 200 other SNLI training samples:

| **Finetuned Set** | **SNLI Test** | **SNLI contra** | **Adv. test** |
|---|---|---|---|
| Baseline | **89.2** | 91.2 | 62.0 |
| 300 SNLI | 89.1 | 90.6 | 62.0 |
| 100 adversarial | 88.4 | **93.7** | 72.0 |
| 100 adversarial + 200 SNLI | 88.9 | 92.9 | **75.0** |

From this we can determine that just training on the 100 new adversarial training set examples is the best for the most similar SNLI contradiction set, but is the worst for the base SNLI test set, and is not the best for our new adversarial test set.

The method we did choose (mixing the 100 new adversarial training examples with 200 original SNLI training examples) was only the best in the adversarial test category, but it performed quite well (2nd) on the most similar SNLI contradictions, and still performed quite good on the original SNLI test set as well.

The major takeaway from this ablation test is that there are tradeoffs to every training setup and dataset, so it's important to deeply understand the problem you are trying to solve, and how this solution may help or hurt in your particular case.

For another ablation test, we got predictions for all three of the discussed test sets with a range of adversarial training set sizes, from 0 to 100 samples of each class:

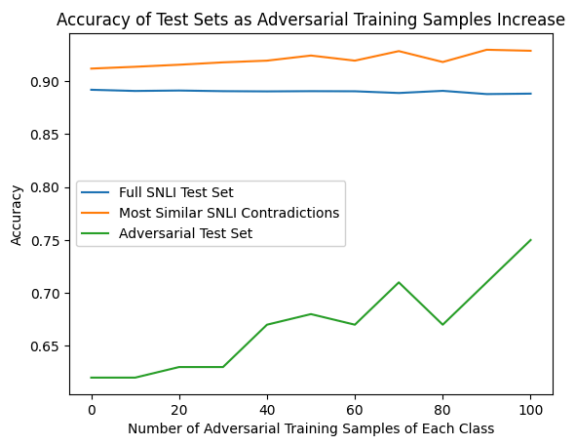

On this chart we can see that as the number of adversarial training samples increase, the full SNLI test set goes down a bit (though stays approximately the same), while the most similar SNLI contradiction set goes up a bit, and the adversarial test set goes up quite a bit, and continues to go up as we add more and more adversarial training samples.

This shows that there is a tradeoff for fixing the problem of similar words and phrases in the premise and hypothesis, and that is that accuracy on the rest of the SNLI test set goes down. However, since the total test set accuracy doesn't decrease as fast as the other sets are increasing, it's possible that further research could find a way to better incorporate the adversarial training set to actually increase the total test set accuracy.

## 6. Conclusions

From our results we can conclude that we successfully identified and fixed a shortcoming of our baseline model for the SNLI task. The addition of finetuning on a relatively small adversarial training set was able to help both the adversarial test set, as well as the underlying SNLI test set for the samples we identified: the most similar contradiction samples.

We weren't able to improve the topline accuracy metric however, which means there is still room for improvement with our fix. If we could find a way to address the problem without degrading the other categories, we would be able to improve both our problem's metric as well as the top line metric.

Based on our graphs, it's also very possible that there is significant room for improvement with a much larger adversarial training set. We were limited in time and scope for this investigation, but a promising future research direction would be to create many more manual adversarial test and training samples and repeat these experiments with those larger datasets.

Overall however, this investigation underscores the importance of deeply understanding the training and test dataset for any machine learning project. It also highlights the possible tradeoffs in natural language tasks, and how different data and training strategies can be used for various outcomes.